\documentclass{article}
\usepackage{ijcai19}
\usepackage{times}
\usepackage{epsfig}
\usepackage{graphicx}
\usepackage{amsmath}
\usepackage{amssymb}
\usepackage{algorithm}
\usepackage{algpseudocode}
\usepackage{subfigure}
\usepackage{multirow}
\usepackage{booktabs}
\usepackage{mathrsfs}
\usepackage{tabu}
\usepackage{float}
\usepackage{subfigure}
\usepackage{multirow}
\usepackage{array}
\usepackage{enumerate}
\usepackage{booktabs}
\usepackage{bm}
\usepackage{mathrsfs}
\usepackage{indentfirst}
\usepackage{times}
\usepackage{soul}
\usepackage{url}
\usepackage[hidelinks]{hyperref}
\usepackage[utf8]{inputenc}
\usepackage[small]{caption}
\usepackage{graphicx}
\usepackage{amsmath}
\usepackage{booktabs}
\title{{RBCN:}  Rectified Binary Convolutional Networks for Enhancing  the \\Performance of 1-bit DCNNs}

\author{
Chunlei Liu$^1$\and
Wenrui Ding$^2$\and
Xin Xia$^1$\and
Yuan Hu$^1$\and
Baochang Zhang$^3$\footnote{Contact Author}\and\\
Jianzhuang Liu$^4$\and
Bohan Zhuang$^5$\and
Guodong Guo$^{6,7}$\\
\affiliations
$^1$ School of Electronic and Information Engineering, Beihang University,\\
$^2$ Unmanned System Research Institute, Beihang University,\\
$^3$ School of Automation Science and Electrical Engineering, Beihang University,\\
$^4$ Huawei Noah's Ark Lab,
$^5$ The University of Adelaide, Australia,\\
$^6$ Institute of Deep Learning, Baidu Research,\\
$^7$ National Engineering Laboratory for Deep Learning Technology and Application\\
\emails
\{liuchunlei, ding, xinxia, zy1602hy, bczhang\}@buaa.edu.cn,
liu.jianzhuang@huawei.com,
bohan.zhuang@adelaide.edu.au,
guoguodong01@baidu.com
}

\begin{document}

\maketitle

\begin{abstract}
Binarized  convolutional neural networks (BCNNs) are widely used to improve memory and computation efficiency of deep convolutional neural networks (DCNNs) for mobile and AI chips based applications.
However, current BCNNs are not able to fully explore their corresponding full-precision models, causing a significant performance gap between them.
In this paper, we propose rectified binary convolutional networks (RBCNs), towards optimized BCNNs, by combining full-precision kernels and feature maps to rectify the binarization process in a unified framework.
In particular, we use a GAN to train the 1-bit binary network with the guidance of its corresponding full-precision model, which significantly improves the performance of BCNNs.
The rectified convolutional layers  are generic and flexible, and can be easily incorporated into  existing DCNNs such as WideResNets and ResNets.
Extensive experiments  demonstrate the superior performance of the proposed RBCNs over state-of-the-art BCNNs.
In particular, our method shows strong generalization on the object tracking task.
\end{abstract}

\section{Introduction}
Deep convolutional neural networks (DCNNs) have been successfully demonstrated on many computer vision tasks such as object detection and image classification.
DCNNs deployed in practical environments, however, still face many challenges.
They  usually involve millions of parameters and  billions of FLOPs during computation.
This is critical because models of vision applications may consume very large amounts of memory and computation, making them impractical for most embedded platforms.
\begin{figure*}[!t]
\centering
\includegraphics[scale=0.255]{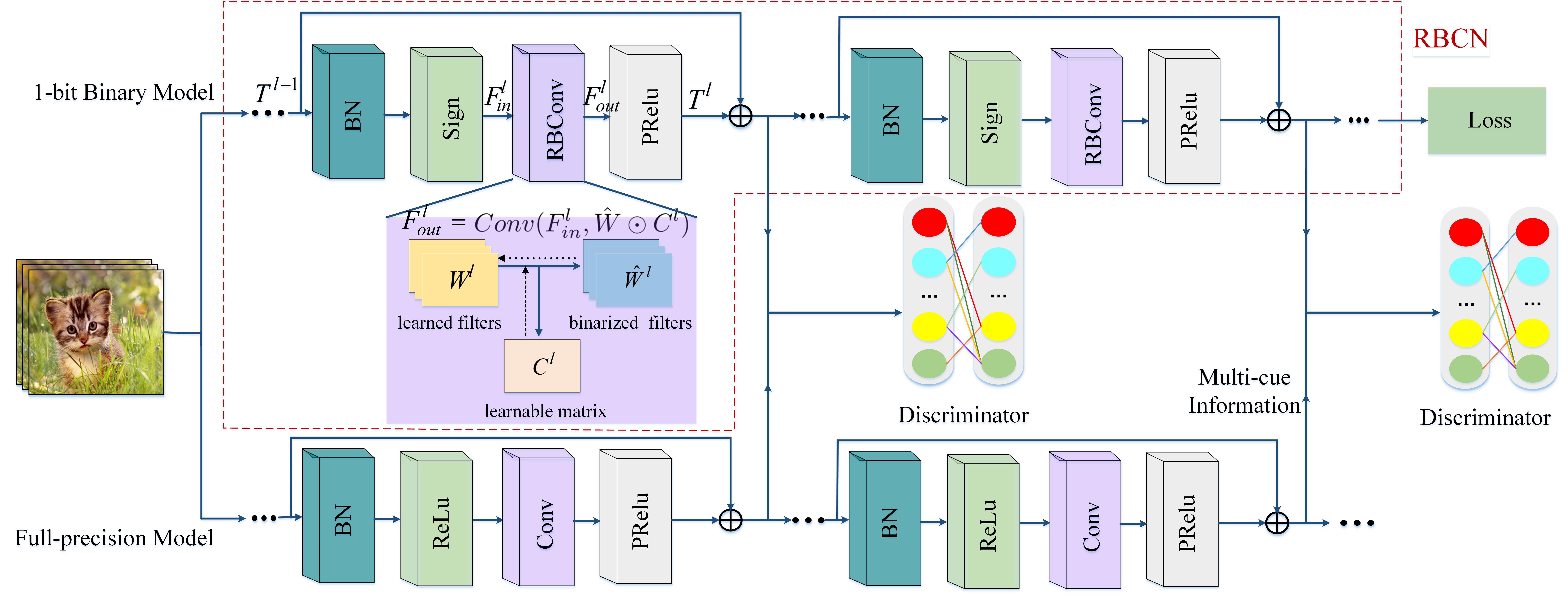}
\caption{The framework of  Rectified Binary Convolutional Network (RBCN). The full-precision model provides the ``real'' feature maps, while the 1-bit model (as the generator) provides the ``fake'' feature maps, to the discriminators that try to distinguish the ``real'' from the ``fake''. Meanwhile, the generator tries to make the discriminators unable to work well. By repeating this process,  both the full-precision feature maps and  kernels (across all the convolutional layers) are sufficiently employed  to enhance the capacity of the 1-bit binary model. Note that (1) the full-precision model is used only in learning but not in inference; (2) after training, the full-precision learned filters $W$ are discarded, and only the binarized filters $\hat{W}$ and the shared  learnable matrixs $C$ ($C^*$) are kept in RBCN for the calculation of the feature maps in inference.}
\label{main}
\end{figure*}

Binary filters instead of using full-precision filter weights have been investigated in DCNNs to compress the deep models to handle the aforementioned problems.
Many works  attempt to quantize the weights of a network while keeping the activations (feature maps) to 32-bit floating points \cite{zhou2017incremental,zhu2016trained,Wang_2018_CVPR}. Although this scheme leads to less performance decrease compared to its full-precision counterpart, it still needs a substantial amount of computational resource to handle the full-precision activations.
Therefore, the so-called 1-bit DCNNs, which target the problem of training the networks with both 1-bit quantized weights and 1-bit activations, become more promising and significant in the field of DCNNs compression.  As presented in \cite{rastegari2016xnor}, by reconstructing the full-precision filters with a single scaling factor, XNOR provides an efficient implementation of convolutional operations.
More recently, Bi-Real Net \cite{liu2018bi} explores a new variant of residual structure to preserve the real activations before the sign function.
And the researchers in \cite{hou2016loss} propose a new value approximation method that considers the effect of binarization on the loss to further obtain binarized weights.  PCNN \cite{Gu2019P} learns a set of diverse quantized kernels by exploiting multiple projections with discrete back propagation.

The investigation into prior arts reveals that how to use the  full-precision models is the key issue to obtain the optimized BCNNs.
Most existing methods use the full-precision models as an initialization \cite{rastegari2016xnor} \cite{liu2018bi}, or for kernel approximation \cite{Gu2019P} \cite{rastegari2016xnor}. Besides, knowledge distillation uses a teacher model (e.g., a full-precision model) to provide a guidance to quantize the network \cite{Polino2018Model,zhuang2018towards,Mishra2017Apprentice}. While these methods generally use a regularization term to  minimize the difference between the student's  and teacher's posterior probabilities or intermediate feature representations, they fail to consider the full-precision feature maps (activations)  in a comprehensive way. This might be the reason why the knowledge distillation methods have not been employed to  obtain the extreme 1-bit CNNs yet.   To narrow down the performance gap between a BCNN and its full-precision model, we propose that  the full-precision kernels and feature maps should be considered in a more comprehensive  way, in order to fully exploit the multi-cue information.

In this paper, we introduce a rectified binary convolutional network (RBCN) to calculate an optimized BCNN in which a novel learning architecture is introduced to combine the full-precision feature maps and the kernels approximation in an end-to-end manner.
Based on the powerful probability fitting ability of generative adversarial network (GAN), we discover that  training a BCNN network with GAN, a better performance can be obtained by fitting the distribution of feature maps between full-precision and 1-bit binary networks.
By doing so, GAN is introduced to distill RBCN from full-precision network by exploiting their full-precision feature maps.
To the best of our knowledge, we are the first to use a GAN to do binary approximation of the full-precision model. The whole process is illustrated in Fig.~\ref{main}, where the full-precision model and the 1-bit binary model (generator) respectively provide ``real'' and ``fake'' feature maps to the discriminators. The discriminators try to distinguish the ``real'' from the ``fake'', and the generator tries to make the discriminators unable to work well. By repeating this process, the multi-cue information (full-precision kernels and feature maps) is sufficiently employed in the training process to enhance the representational ability of the 1-bit binary model. Besides, kernel (filter) approximation (RBConv in Fig.~\ref{main}) is integrated in the framework. Also, multiple discriminators are used to further improve the performance of RBCN. This process involving the GAN and the kernel approximation is a rectified process, which can lead to a unique architecture with more precise estimation of the full-precision model. The contributions of this paper are summarized as follows.

(1) A novel BCNN learning architecture, referred to as rectified binary convolutional network (RBCN), is proposed, which employs the full-precision kernels and feature maps to rectify the binarization process in a comprehensive framework.


(2) To the best of our knowledge, we are the first to use a GAN to calculate a BCNN. Besides, we discover that using multiple discriminators in the GAN can significantly improve the performance of the 1-bit binary model.

(3) Extensive experiments  demonstrate the superior performance of the proposed RBCNs over state-of-the-art BCNNs on the object classification and tracking tasks.

\begin{table*}[]
\centering
\caption{A brief description of the variables and operators used in the paper.}
\begin{tabular}{llllllll}
\hline
 $\mathscr{L}:$ loss function &$\hat{W}:$ binarized filters &$T:$ feature maps from RBCN to $D(\cdot)$ &$D(\cdot):$ discriminators    &\\
$W:$ learned filters &$C:$   learnable matrixs& $R:$ feature maps from the full-precision model& ${\delta}_C:$  gradient of  $C$&\\\hline
{$i:$ filter index}     & ${\eta}:$ learning rate   &$F:$ feature maps before and after convolution in RBCN&$l:$ layer index&   \\
$t:$ $t^{th}$ iteration & $L:$ number of layers  &$Y:$ filters of the discriminators&${\delta}_W:$  gradient of  $W$ &     \\\hline
\end{tabular}
\label{notation}
\end{table*}

\section{Rectified Binary Convolutional Networks (RBCNs)}
We design RBCNs via kernel approximation and training with GANs to rectify BCNNs in a unified framework.
During this process, the multi-cue information of the full-precision feature maps and kernels\footnote{In this paper, the terms ``filter'' and ``kernel'' are exchangeable.} is exploited to improve the performance degraded by binarization.
The rectified convolutional layers are generic and flexible, which can be easily incorporated into existing CNNs, such as WideResNets and ResNets.
First of all, Table 1 gives the main notation used in this paper.

\subsection{Loss Function of RBCNs}
The rectified process combines the full-precision kernels and feature maps to rectify the binarization process. It includes kernel approximation and adversarial learning.
This learnable kernel approximation can lead to an unique architecture with more precise estimation of the original convolutional filters through minimizing a kernel loss.
The discriminators $D(\cdot)$ with filters $Y$ are introduced to distinguish the feature maps $R$ of the full-precision model from those $T$ of RBCN.
The generator (RBCN) with filters $W$ and learnable matrixs $C$ is learned together with $Y$ by using the knowledge from the supervised feature maps $R$.
Therefore, $W$, $C$ and $Y$ are learned by solving the following optimization problem:
\begin{small}
\begin{equation}
\begin{split}
\begin{matrix}
   \begin{matrix}
   \arg\min\limits_{W,\hat{W},C}\max\limits_{Y}\!&\mathscr{L}=\mathscr{L}_{Adv}(W,\hat{W},C,Y)\qquad\qquad\qquad\qquad\\
   &+\mathscr{L}_{Kernel}(W,\hat{W},C)+\mathscr{L}_{S}(W,\hat{W},C),
\end{matrix}
\end{matrix}
\end{split}
\label{optimization1}
\end{equation}\end{small}

\noindent where $\mathscr{L}_{Adv}(W,\hat{W},C,Y)$ is the adversarial loss:
\begin{small}
\begin{equation}
\begin{split}
{\mathscr{L}_{Adv}(W,\hat{W},C,Y)=log(D(R;Y))+log({1-D(T;Y)})},
\end{split}
\label{optimization2}
\end{equation}\end{small}
\begin{flushleft}where $D(\cdot)$ consists of four basic blocks, each of which has a linear layer and a LeakyRelu layer.
\end{flushleft}

In addition, $\mathscr{L}_{Kernel}(W,\hat{W},C)$ is the kernel loss between the learned full-precision filters $W$ and the binarized filters $\hat{W}$, which is expressed by MSE:
\begin{equation}
\begin{split}
   \mathscr{L}_{Kernel}(W,\hat{W},C)=\frac{1}{2}\lambda_1||W-C\hat{W}|{{|}^{2}}.
\label{optimization3}
\end{split}
\end{equation}
Finally, $\mathscr{L}_{S}(W,\hat{W},C)$ is a traditional problem-dependent loss such as the softmax loss.

For simplicity, the update of the discriminators is omitted in the following description until Algorithm \ref{algo}.
Besides, we find that the $log$ in Equ.~\ref{optimization2} has little effect during training and so it is omitted  too.
 Then, based on the Lagrangian method, the optimization problem in Equ.~\ref{optimization1} is rewritten as:
\begin{equation}
\begin{split}
   & \min \qquad \mathscr{L}_S(W,\hat{W},C)\\
   &+\lambda_1/2  \sum_{l} \sum_{i} ||W_i^{l}-C^{l}\hat{W}_i^{l}|{{|}^{2}}\\
   &+\lambda_2/2\sum_{l}\sum_{i}||{1-D(T_i^{l};Y)}||^2.
\label{optimization4}
\end{split}
\end{equation}
In Equ.~\ref{optimization4}, the target is to obtain $W$, $\hat{W}$ and $C$ with $Y$ fixed, which is why the term $D(R;Y)$ in Equ.~\ref{optimization2} can be ignored.
The update of $Y$ can be found in Algorithm \ref{algo}.
The advantage of our formulation in Equ.~\ref{optimization4} lies in that the loss function is trainable, meaning that it can be easily incorporated into existing learning frameworks.

\subsection{Forward Propagation in RBCNs}
In RBCNs, a binary filter $\hat{W}_i^l$ is calculated as:
\begin{equation}
\hat{W_i^l} = {\rm{sign}}(W_i^l),
\label{M}
\end{equation}
where $W_i^l$ is the corresponding full-precision filter, and the values of $\hat{W}_i^l$ are $1$ or $-1$. Both $W_i^l$ and $\hat{W}_i^l$ are jointly obtained in the end-to-end learning.

In RBCNs, the convolution is implemented based on $C^l$ and $F_{in}^l$ to calculate the feature maps $F_{out}^{l}$:
\begin{equation}
\begin{split}
F_{out}^{l} &= RBConv(F_{in}^l; \hat{W}^l, C^l)\\
&=Conv(F_{in}^l,\hat{W}\odot{C^l}),
\label{5}
\end{split}
\end{equation}
where $RBConv$ denotes the convolution operation implemented as a new module, $F_{in}^l$ and $F_{out}^{l}$  are the feature maps before and after the convolution, respectively, and $\odot$ is the element-by-element product. Note that $F_{in}^l$ is binary after the sign operation (see Fig.~\ref{main}), and $C$ is actually $C^*$, which will be elaborated at the end of section 3.3.
\subsection{Backward Propagation in RBCNs}
In RBCNs, what need to be learned and updated are the full-precision filters $W$ and the learnable matrixs $C$.
These two sets of parameters are jointly learned.
In each convolutional layer, an RBCN updates $W$ first and then $C$.

\subsubsection{Updating $W$}
Let ${{\delta }_{W_i^l}}$ be the gradient of the full-precision filter $W_i^l$. During backpropagation, the gradients pass to $\hat{W}_i^l$ first and then to $W_i^l$. Thus:
\begin{equation}
{{\delta }_{W_i^l}}=\frac{\partial \mathscr{L}}{\partial W_i^l}=\frac{\partial \mathscr{L}}{\partial \hat{W_i^l}}\frac{\partial \hat{W_i^l}}{\partial W_i^l},
\label{7}
\end{equation}
where
\begin{equation}
\frac{\partial \hat{W}_i^l}{\partial W_i^l}=\left\{ \begin{matrix}
   2+2W_i^l,\qquad-1\le W_i^l<0,  \\
   2-2W_i^l,\quad\quad0\le W_i^l<1,  \\
   \begin{matrix}
   0,\qquad\qquad\quad{\rm{otherwise}}, & {}  \\
\end{matrix}  \\
\end{matrix} \right.
\label{8}
\end{equation}
which is an approximation of the $2\times$dirac-delta function \cite{liu2018bi}. Furthermore,
\begin{equation}
\frac{\partial \mathscr{L}}{\partial {\hat{W}_i^l}}=\frac{\partial {\mathscr{L}_{S}}}{\partial {\hat{W_i^l}}}+\frac{\partial {\mathscr{L}_{Kernel}}}{\partial {\hat{W_i^l}}}+\frac{\partial {\mathscr{L}_{Adv}}}{\partial {\hat{W_i^l}}},
\end{equation}
and
\begin{equation}
{{W_i^l}}\leftarrow {{W_i^l}}-\eta_1 {{\delta }_{W_i^l}},
\end{equation}
where $\eta_1$ is a learning rate. Then:
\begin{equation}
\begin{split}
\frac{\partial {\mathscr{L}_{Kernel}}}{\partial {\hat{W}_i^{l}}}&=-\lambda_1 ({W^{l}_i-C^{l}\hat{W}^{l}_i})C^{l},
\end{split}
\end{equation}
\begin{equation}
\frac{\partial {\mathscr{L}_{Adv}}}{\partial {\hat{W}_i^{l}}}=-\lambda_2 ({1-D(T_i^{l};Y)})\frac{\partial {D}}{\partial {{\hat{W}^{l}}}}.
\end{equation}

\subsubsection{Updating $C$}
We further update the learnable matrix $C^l$ with $W^l$ fixed. Let ${{\delta }_{C^l}}$ be the gradient of $C^l$. Then we have:
\begin{equation}
{{\delta }_{C^l}}=\frac{\partial \mathscr{L}_S}{\partial C^l}+\frac{\partial \mathscr{L}_M}{\partial {C^l}}+\frac{\partial \mathscr{L}_{Adv}}{\partial {C^l}},
\label{13}
\end{equation}
and
\begin{equation}
{{C^{l}}}\leftarrow {{C^{l}}}-\eta_2 {{\delta }_{C^{l}}},
\end{equation}
where $\eta_2$ is another learning rate.
Further,
\begin{equation}
\begin{split}
\frac{\partial {\mathscr{L}_{Kernel}}}{\partial {{C^{l}}}}&=-\lambda_1\sum_{i}({W^{l}_i-C^l\hat{W}^{l}_i}) \hat{W}^{l}_i,
\end{split}
\end{equation}
\begin{equation}
\frac{\partial {\mathscr{L}_{Adv}}}{\partial {{C^{l}}}}=-\lambda_2\sum_{i} ({1-D(T^l_i;Y)})\frac{\partial {D}}{\partial {{C^{l}}}}.
\end{equation}

The above derivations show that the rectified process is trainable in an end-to-end manner. The complete training process is summarized in Algorithm \ref{algo}, including the update of the discriminators. Besides, in the implementation, the batch normalization (BN) layers are updated with $W$ and $C$ fixed after each epoch.

We note that in our implementation, the value of  $C$  will be replaced by its average  during the forward process, resulting into a new matrix denoted by $C^*$\footnote{its elements are equal}. By doing so, only a scalar instead of a matrix involve into the convolution  which thus speed up the calculation.

\begin{algorithm}[htb]
\caption{ RBCN  Training}
\label{algo}
\begin{algorithmic}[1]
\Require
 The training dataset, the feature maps $R$ from the full-precision model,
 and the hyper-parameters such as initial learning rate, weight decay,
 convolution stride and padding size.
\Ensure
 A binary 1-bit model RBCN with weights $\hat{W}$ and learnable matrixs $C$.
\State Initialize $W$ randomly;
\Repeat
\State Randomly sample a mini-batch data;
\State // Forward propagation
\ForAll {$l = 1$ to $L$ convolutional layer}
\State $F_{out}^{l} = Conv(F_{in}^l,\hat{W}^l\odot C^l)$;
\EndFor
\State // Back propagation
\ForAll {$l = L$ to $1$}
\State Update the discriminators $D^l(\cdot)$ by ascending  their stochastic gradients:
\State \qquad\begin{small}$ \nabla_{D^l}(log(D^l(R^l;Y))+log({1-D^l(T^l;Y)}));$\end{small}
\State Calculate the gradients ${\delta}_{W^l}$; //\,\,Using Eq.\;\ref{7}
\State  $W^l \leftarrow W^l - {\eta} {\delta}_{W^l}$; //\,\, Update the weights
\State Calculate the gradient ${\delta}_{C^l}$; //\,\,Using Eq.\;\ref{13}
\State  ${{C^{l}}}\leftarrow {{C^{l}}}-\eta_2 {{\delta }_{C^{l}}}$; //\,\,Update the learnable matrixs
\EndFor
\State Update all the parameters of the batch normalization layers
\Until{the maximum epoch}
\State $\hat{W}={\rm{sign}}(W)$.
\end{algorithmic}
\end{algorithm}

\section{Experiments}
Our RBCNs are evaluated first on  object classification  using MNIST \cite{L1998Gradient}, CIFAR10/100 \cite{Krizhevsky2009Learning} and ILSVRC12 ImageNet datasets \cite{Russakovsky2015ImageNet}, and then on object tracking. For object classification, WideResNet (WRN) \cite{Zagoruyko2016Wide} and  ResNet \cite{He2016Deep}  are employed as the backbone networks to build our RBCNs. Also,  binarizing the neuron activations is carried out in all of our experiments.
\subsection{Datasets and Implementation Details}
\textbf{Datasets:}  The MINIST \cite{L1998Gradient} dataset is composed of a training set of 60,000 and a testing set of 10,000 $32\times 32$ grayscale images of hand-written digits from 0 to 9.

CIFAR10 \cite{Krizhevsky2009Learning} is a natural image classification dataset containing a training set of $50,000$ and a testing set of $10,000$ $32\times 32$ color images across the following 10 classes: airplanes, automobiles, birds, cats, deers, dogs,
frogs, horses, ships, and trucks, while CIFAR100 consists of 100 classes.

ImageNet object classification dataset \cite{Russakovsky2015ImageNet} is more challenging due to its large scale and greater diversity.
There are 1000 classes and 1.2 million training images and 50k validation images in it. We compare our method with the state-of-the-art on the ImageNet dataset, and we adopt ResNet18 to validate the superiority and effectiveness of RBCNs.

\textbf{WRN Backbone:} WRN is a network structure similar to ResNet with a depth factor $k$ to control the feature map depth dimension expansion through 3 stages, within which the dimensions remain unchanged.
For simplicity we fix the depth factor to 1.
Each WRN has a parameter $i$ which indicates the channel dimension of the first stage, and we set it to 16, leading to a network structures $16$-$16$-$32$-$64$. The training details are the same as in \cite{Zagoruyko2016Wide}.
$\lambda_1$ and $\lambda_2$ are set as 0.01 with a degradation of 10\% for every 60 epochs before reaching the maximum epoch of 200 for CIFAR10/100.
For example, WRN22 is a network with 22 convolutional layers and similarly for WRN40.

\textbf{ResNet18 Backbone:}
SGD is used as the optimization algorithm with a momentum of $0.9$ and a weight decay 1e-4.
$\lambda_1$ and $\lambda_2$ are set as  0.1 with a degradation of 10\% for every 20 epochs before reaching the maximum epoch of 70 on ImageNet, while on CIFAR10/100, $\lambda_1$ and $\lambda_2$ are set as 0.01 with a degradation of 10\% for every 60 epochs before reaching the maximum epoch of 200.

\subsection{Ablation Study}
In this section, we study the performance contributions of the components in RBCNs, which include kernel approximation, GAN, and the update of the BN layers. CIFAR100 and ResNet18 with different kernel stages are used in this experiment.
The details are given below.

\begin{table}[]
\centering
\caption{Performance (accuracy, \%) contributions of the components in RBCNs on CIFAR100, where Bi, R, G, and B denote the Bi-Real Net, $RBConv$, GAN, update of the BN layers, respectively. The bold numbers represent the best results.}
\begin{tabular}{cccccc}
\hline
&Kernel Stage & Bi & R & R+G & R+G+B \\\hline\hline
RBCN  &32-32-64-128 & 54.92&56.54 & 59.13& \textbf{61.64}\\
RBCN  &32-64-128-256 &63.11 &63.49 &64.93  &\textbf{65.38}\\
RBCN  &64-64-128-256 & 63.81 &64.13 & 65.02 &\textbf{66.27}\\\hline
\label{ablation}
\end{tabular}
\end{table}

1) We only replace the convolution in Bi-Real Net with our kernel approximation ($RBConv$) and compare the results. As
shown in the R column in Table \ref{ablation}, RBCN  achieves 1.62\% accuracy improvement over Bi-Real Net (56.54\% vs.~54.92\%) using the same network structure as in ResNet18 with 32-32-64-128.~This significant improvement verifies the effectiveness of the learnable matrixs.

2) In RBCNs, if we use the GAN to help binarization, we can find a more significant improvement from 56.54\% to 59.13\% with the kernel stage of 32-32-64-128, which shows that the GAN can really enhance the binarized networks.

3) We find that a training trick can also improve RBCNs, which is to update the BN layers with $W$ and $C$ fixed after each epoch (line 17 in Algorithm \ref{algo}). This trick makes RBCN boost 2.51\% (61.64\% vs.~59.13\%) in CIFAR100 with 32-32-64-128.

\subsection{Accuracy Comparison with  State-of-the-Art}
\begin{table}[]
\centering
\caption{ Classification accuracy (\%) based on ResNet18 and WRN40 on CIFAR10/100. The bold represent the best results among the binary networks. }
\begin{tabular}{cccc}
\hline
\multirow{3}{*}{Model}&\multirow{3}{*}{Kernel Stage}& \multicolumn{2}{c}{Dataset} \\ \cline{3-4}
                      & & CIFAR     & CIFAR   \\
                      & & -10&-100 \\ \hline\hline
ResNet18             &32-32-64-128 & 92.67        & 67.07        \\
ResNet18                  &32-64-128-256 & 93.88         & 72.51        \\
ResNet18             &64-64-128-256 & 94.57        & 72.89        \\
RBCN (ResNet18)                  &32-32-64-128 & 89.03         & \textbf{61.09}        \\
RBCN (ResNet18)                  &32-64-128-256 &   90.67       &   \textbf{65.38}      \\
RBCN (ResNet18)                  &64-64-128-256 &   90.40       &   \textbf{66.27}      \\\hline
WRN22                &64-64-128-256  & 95.19         & 76.38        \\
WRN40                &64-64-128-256  & 94.92         & 74.91        \\
RBCN (WRN22)                 &64-64-128-256&  93.28       &  \textbf{72.06}       \\
RBCN (WRN40)                 &64-64-128-256&  93.69       &  \textbf{73.08}      \\\hline
XNOR (ResNet18)                  &32-32-64-128&   71.01     &  43.58       \\
XNOR (WRN22)                  &64-64-128-256 &    86.90    &     58.05    \\
Bi-Real (ResNet18)                  &32-32-64-128&   85.34      &   54.92      \\
Bi-Real (WRN22)                  &64-64-128-256 &   90.65      &  68.51       \\
PCNN (ResNet18)                  &32-32-64-128&     85.50    &   55.66      \\
PCNN (WRN22)                  &64-64-128-256 &    91.62     &     70.32    \\
Scheme-A (ResNet18)                  &32-64-128-256 &  75.45      &   46.32      \\
Scheme-A (WRN22)                  &64-64-128-256 &    87.83     &    59.54     \\\hline
\end{tabular}
\label{cifartable}
\end{table}
\textbf{CIFAR10/100:} The same parameter settings are used in RBCNs  on both CIFAR10 and CIFAR100.
We first compare our RBCNs with the original ResNet18 with different stage kernels,
followed by a comparison with the original WRNs with the initial channel dimension $64$ in Table \ref{cifartable}.
Thanks to the rectified process, our results on both the datasets are close to the full-precision networks ResNe18 and WRN40.
Then, we compare our results with other state-of-the-arts such as Bi-Real Net \cite{liu2018bi}, PCNN \cite{Gu2019P}, Scheme-A \cite{Mishra2017Apprentice} and XNOR \cite{rastegari2016xnor}.
All these BCNNs have both binary filters and binary activations.
It is observed that at most 6.17\% ($=$ 61.09\%$-$54.92\%) accuracy improvement is gained with our RBCN, and in other cases, larger margins are achieved.

\begin{figure*}[!t]
\centering
\includegraphics[scale=0.32]{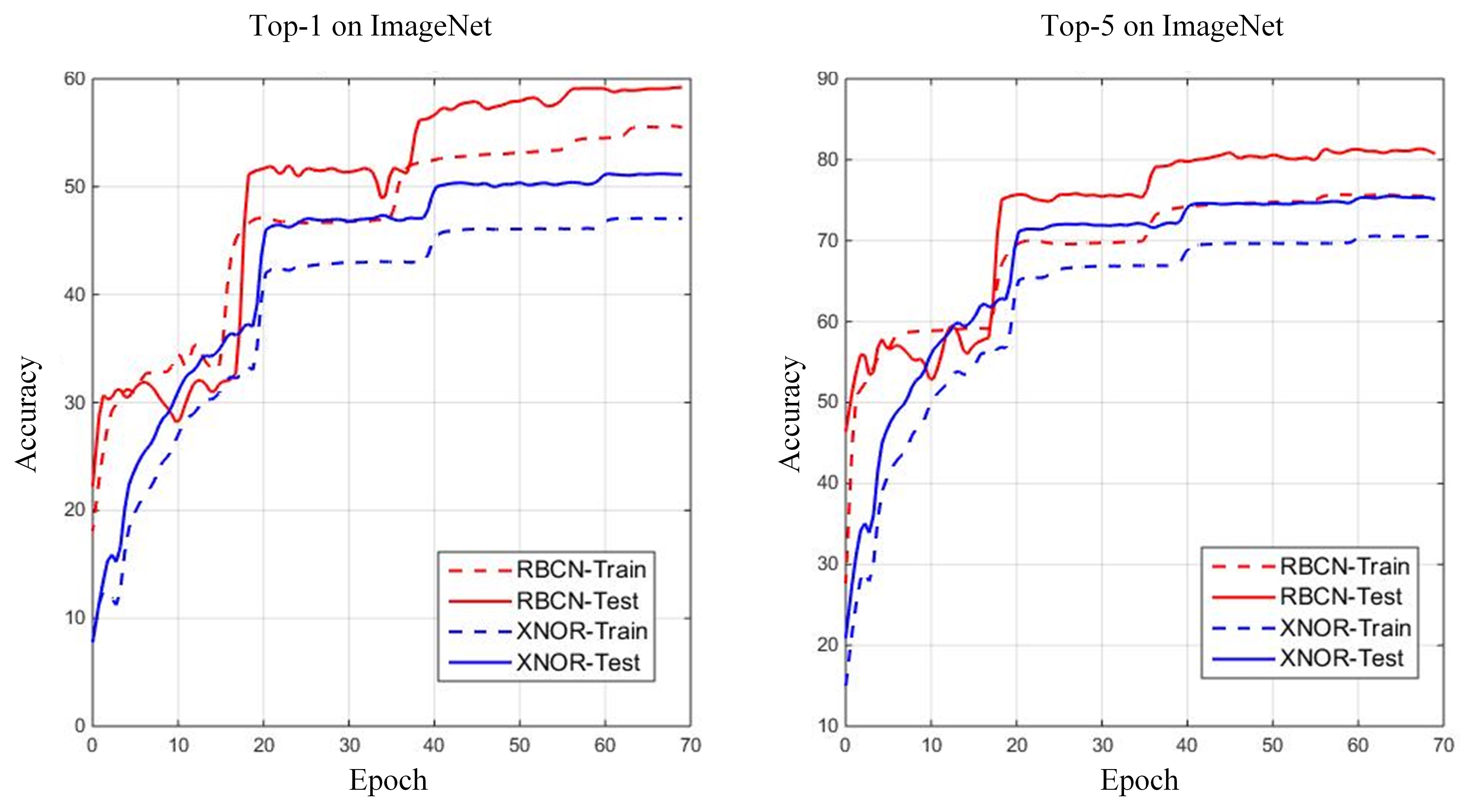}
\caption{Training and Testing error curves of RBCN and XNOR based on the ResNet18 backbone on ImageNet.}
\label{lossimagenet}
\end{figure*}

\textbf{ImageNet:} Five  state-of-the-art methods  on ImageNet are chosen for comparison:
Bi-Real Net \cite{liu2018bi}, BinaryNet \cite{Courbariaux2016Binarized}, XNOR \cite{rastegari2016xnor}, PCNN \cite{Gu2019P} and ABC-Net \cite{lin2017towards}.
Again, these networks are representative methods of binarizing both network weights and activations and achieve state-of-the-art results. All the methods in Table \ref{imagenettable} perform the binarization of ResNet18.
The results in Table \ref{imagenettable} are quoted directly from their papers, except that the result of BinaryNet is from \cite{lin2017towards}.
The comparison clearly indicates that the proposed RBCN outperforms the five binary networks by a considerable margin in terms
of both the top-1 and top-5 accuracies.
Specifically, for top-1 accuracy, RBCN outperforms BinaryNet and ABC-Net with a gap over 16\%, achieves 7.9\%  improvement over XNOR, 3.1\% over the very recent Bi-Real Net, and 2.2\% over the latest PCNN.
In Fig. \ref{lossimagenet}, we plot the training and testing loss curves of XNOR and RBCN.
It clearly shows that using our rectified process, RBCN converges faster than XNOR.

\begin{table*}[]
\centering
\caption{ Classification accuracy (\%) on ImageNet. The bold  represents the best result among the binary networks.}
\begin{tabular}{ccccccccc}
\hline
                          &       & Full-Precision & XNOR & ABC-Net & BinaryNet & Bi-Real&PCNN & RBCN      \\ \hline\hline
\multirow{2}{*}{ResNet18} & Top-1 & 69.3         & 51.2   & 42.7  & 42.2    & 56.4 &57.3 & \textbf{59.5} \\
                          & Top-5 & 89.2         & 73.2   & 67.6  & 67.1    & 79.5 &79.8 & \textbf{81.6} \\ \hline
\end{tabular}
\label{imagenettable}
\end{table*}
\begin{table}[]
\centering
\caption{Tracking performance comparison between XNOR and RB-SF on different datasets.}
\begin{tabular}{cccccc}
\hline
  Dataset&Index& SiamFC & XNOR   & RB-SF  \\ \hline
\multirow{2}{*}{GOT-10K}&AO & 0.348  & 0.251 & {0.327} \\ \cline{2-5}
&SR & 0.383  & 0.230  & {0.343} \\ \hline\hline
\multirow{2}{*}{OTB50}&Precision &0.761 &0.457& {0.706}\\ \cline{2-5}
&SR &0.556 &0.323& {0.496}\\ \hline\hline
\multirow{2}{*}{OTB100}&Precision &0.808 &0.541&{0.786} \\ \cline{2-5}
&SR & 0.602 &0.394&{0.572} \\ \hline\hline
\multirow{2}{*}{UAV123}&Precision &0.745 &0.547 &{0.688} \\ \cline{2-5}
&SR &0.528 &0.374 &{0.497} \\ \hline
\label{tracking}
\end{tabular}
\end{table}

\subsection{Experiments on object tracking}
The key message conveyed in the proposed method is that although the conventional binary training method has a limited
model capability, the proposed rectified process can lead to a powerful model.
In this section, we show that this framework can also be used in object tracking.
In particular, we consider the problem of tracking an arbitrary object in videos, where the object is identified solely by a rectangle in the first frame.
For object tracking, it is necessary to update the weights of the network online, severely compromising the speed of the system.
To directly apply the proposed framework to this application, we can construct a binary convolution with the same structure to reduce the convolution time.
In this way, RBCN can be used to binarize the network further to guarantee the tracking performance.

In this paper, we use SiamFC Network as the backbone for object tracking. We binarize SiamFC as Rectified Binary Convolutional SiamFC Network (RB-SF). We evaluate RB-SF on four datasets, GOT-10K \cite{huang2018got}, OTB50 \cite{wu2013online}, OTB100 \cite{wu2015object}, and UAV123 \cite{mueller2016benchmark}, using accuracy occupy (AO) and success rate (SR).
The results are shown in Table \ref{tracking}.
Intriguingly, our framework achieves about 7\% AO improvement over XNOR, both using the same network architecture as in SiamFC Network on GOT-10k. Further, our framework brings so much benefit that Bi-SF performs almost as well as the full-precision SiamFC Network.

\subsection{Efficiency Analysis}

The memory usage is computed as the summation of 32 bits times the number of real-valued parameters and 1 bit times the number of binary parameters in the network.
Further, we use FLOPs to measure the speed. The results are given in Table \ref{Efficiency Analysis}. The FLOPs are calculated as the amount of real-valued floating point multiplications plus 1/64 of the amount of 1-bit multiplications \cite{liu2018bi}.
As shown in Table \ref{Efficiency Analysis}, the proposed RBCN, along with XNOR, reduces the memory usage of the full-precision ResNet18 by 11.10 times.
For efficiency, both RBCN and XNOR gain $10.86\times$ speedup over ResNet18.
Note the computational and storage costs brought by learnable scalar $C^*$ can be negligible.
\begin{table}[]
\centering
\caption{Comparison of memory usage and FLOPs calculation.}
\begin{tabular}{cccc}
\hline
              & RBCN & XNOR-Net &{ResNet18} \\ \hline
Memory usage  & $33.7$Mbits     &    $33.7$Mbits      & $374.1$Mbits         \\ \hline
Memory saving & $11.10 \times$     &    $11.10 \times$       &-         \\ \hline
FLOPs         &  $1.67\times10^8$     &  $1.67\times10^8$        & $1.81\times10^9$         \\ \hline
Speedup       &   $10.86 \times$   &   $10.86 \times$        & -         \\ \hline
\end{tabular}
\label{Efficiency Analysis}
\end{table}
\section{Conclusion}

In this paper, we introduce rectified binary convolutional networks (RBCNs), towards optimized BCNNs, by exploiting the full-precision kernels and feature maps in an end-to-end manner.
In particular, we use a GAN to train the 1-bit binary network with the guidance of its corresponding full-precision model, which significantly improves the performance of the BCNN.
Furthermore, as a general model, RBCNs can be used not only in object classification but also in other tasks such as object tracking.
The experiments on both object classification and object tracking demonstrate the superior performance of the proposed RBCNs over state-of-the-art binary models.

\section{Acknowledgment}
The work was supported by the National Key Research and Development Program of China
(Grant No. 2016YFB0502602) and the Natural Science Foundation of China under Contract 61672079. Also, it is in part supported by the Fundamental Research Funds for the Central Universities. Baochang Zhang is the corresponding author.

\bibliographystyle{named}
\bibliography{egbib}

\begin{thebibliography}{}

\bibitem[\protect\citeauthoryear{Courbariaux \bgroup \em et al.\egroup
  }{2016}]{Courbariaux2016Binarized}
Matthieu Courbariaux, Itay Hubara, Daniel Soudry, Ran El-Yaniv, and Yoshua
  Bengio.
\newblock Binarized neural networks: Training deep neural networks with weights
  and activations constrained to +1 or -1.
\newblock {\em arXiv preprint arXiv:1602.02830}, 2016.

\bibitem[\protect\citeauthoryear{Gu \bgroup \em et al.\egroup }{2019}]{Gu2019P}
Jiaxin Gu, Baochang Zhang, and Jianzhuang Liu.
\newblock Projection convolutional neural networks.
\newblock In {\em AAAI}, 2019.

\bibitem[\protect\citeauthoryear{He \bgroup \em et al.\egroup
  }{2016}]{He2016Deep}
Kaiming He, Xiangyu Zhang, Shaoqing Ren, and Jian Sun.
\newblock Deep residual learning for image recognition.
\newblock In {\em IEEE Conference on Computer Vision and Pattern Recognition},
  pages 770--778, 2016.

\bibitem[\protect\citeauthoryear{Hou \bgroup \em et al.\egroup
  }{2016}]{hou2016loss}
Lu~Hou, Quanming Yao, and James~T Kwok.
\newblock Loss-aware binarization of deep networks.
\newblock {\em arXiv preprint arXiv:1611.01600}, 2016.

\bibitem[\protect\citeauthoryear{Huang \bgroup \em et al.\egroup
  }{2018}]{huang2018got}
Lianghua Huang, Xin Zhao, and Kaiqi Huang.
\newblock Got-10k: A large high-diversity benchmark for generic object tracking
  in the wild.
\newblock {\em arXiv preprint arXiv:1810.11981}, 2018.

\bibitem[\protect\citeauthoryear{Krizhevsky and
  Hinton}{2009}]{Krizhevsky2009Learning}
Nair Krizhevsky and Hinton.
\newblock The cifar-10 dataset.
\newblock {\em online: http://www. cs. toronto. edu/kriz/cifar. html}, 2009.

\bibitem[\protect\citeauthoryear{Lecun \bgroup \em et al.\egroup
  }{1998}]{L1998Gradient}
Yann Lecun, Leon Bottou, Yoshua Bengio, and Patrick Haffner.
\newblock Gradient-based learning applied to document recognition.
\newblock {\em Proceedings of the IEEE}, 86(11):2278--2324, 1998.

\bibitem[\protect\citeauthoryear{Lin \bgroup \em et al.\egroup
  }{2017}]{lin2017towards}
Xiaofan Lin, Cong Zhao, and Wei Pan.
\newblock Towards accurate binary convolutional neural network.
\newblock In {\em Advances in Neural Information Processing Systems}, pages
  345--353, 2017.

\bibitem[\protect\citeauthoryear{Liu \bgroup \em et al.\egroup
  }{2018}]{liu2018bi}
Zechun Liu, Baoyuan Wu, Wenhan Luo, Xin Yang, Wei Liu, and Kwang-Ting Cheng.
\newblock Bi-real net: Enhancing the performance of 1-bit cnns with improved
  representational capability and advanced training algorithm.
\newblock In {\em Proceedings of the European Conference on Computer Vision},
  pages 722--737, 2018.

\bibitem[\protect\citeauthoryear{Mishra and Marr}{2017}]{Mishra2017Apprentice}
Asit Mishra and Debbie Marr.
\newblock Apprentice: Using~knowledge~distillation techniques to improve
  low-precision network accuracy.
\newblock {\em arXiv:1711.05852v1}, 2017.

\bibitem[\protect\citeauthoryear{Mueller \bgroup \em et al.\egroup
  }{2016}]{mueller2016benchmark}
Matthias Mueller, Neil Smith, and Bernard Ghanem.
\newblock A benchmark and simulator for uav tracking.
\newblock In {\em European conference on computer vision}, pages 445--461.
  Springer, 2016.

\bibitem[\protect\citeauthoryear{Polino \bgroup \em et al.\egroup
  }{2018}]{Polino2018Model}
Antonio Polino, Razvan Pascanu, and Alistarh Dan.
\newblock Model compression via distillation and quantization.
\newblock {\em arXiv:1802.05668v1}, 2018.

\bibitem[\protect\citeauthoryear{Rastegari \bgroup \em et al.\egroup
  }{2016}]{rastegari2016xnor}
Mohammad Rastegari, Vicente Ordonez, Joseph Redmon, and Ali Farhadi.
\newblock Xnor-net: Imagenet classification using binary convolutional neural
  networks.
\newblock In {\em European Conference on Computer Vision}, pages 525--542,
  2016.

\bibitem[\protect\citeauthoryear{Russakovsky \bgroup \em et al.\egroup
  }{2015}]{Russakovsky2015ImageNet}
Olga Russakovsky, Jia Deng, Hao Su, Jonathan Krause, Sanjeev Satheesh, Sean Ma,
  Zhiheng Huang, Andrej Karpathy, Aditya Khosla, and Michael Bernstein.
\newblock Imagenet large scale visual recognition challenge.
\newblock {\em International Journal of Computer Vision}, 115(3):211--252,
  2015.

\bibitem[\protect\citeauthoryear{Wang \bgroup \em et al.\egroup
  }{2018}]{Wang_2018_CVPR}
Xiaodi Wang, Baochang Zhang, Ce~Li, Rongrong Ji, Jungong Han, Xianbin Cao, and
  Jianzhuang Liu.
\newblock Modulated convolutional networks.
\newblock In {\em The IEEE Conference on Computer Vision and Pattern
  Recognition}, June 2018.

\bibitem[\protect\citeauthoryear{Wu \bgroup \em et al.\egroup
  }{2013}]{wu2013online}
Yi~Wu, Jongwoo Lim, and Ming-Hsuan Yang.
\newblock Online object tracking: A benchmark.
\newblock In {\em Proceedings of the IEEE conference on computer vision and
  pattern recognition}, pages 2411--2418, 2013.

\bibitem[\protect\citeauthoryear{Wu \bgroup \em et al.\egroup
  }{2015}]{wu2015object}
Yi~Wu, Jongwoo Lim, and Ming-Hsuan Yang.
\newblock Object tracking benchmark.
\newblock {\em IEEE Transactions on Pattern Analysis and Machine Intelligence},
  37(9):1834--1848, 2015.

\bibitem[\protect\citeauthoryear{Zagoruyko and
  Komodakis}{2016}]{Zagoruyko2016Wide}
Sergey Zagoruyko and Nikos Komodakis.
\newblock Wide residual networks.
\newblock {\em arXiv preprint arXiv:1605.07146}, 2016.

\bibitem[\protect\citeauthoryear{Zhou \bgroup \em et al.\egroup
  }{2017}]{zhou2017incremental}
Aojun Zhou, Anbang Yao, Yiwen Guo, Lin Xu, and Yurong Chen.
\newblock Incremental network quantization: Towards lossless cnns with
  low-precision weights.
\newblock {\em arXiv preprint arXiv:1702.03044}, 2017.

\bibitem[\protect\citeauthoryear{Zhu \bgroup \em et al.\egroup
  }{2016}]{zhu2016trained}
Chenzhuo Zhu, Song Han, Huizi Mao, and William~J Dally.
\newblock Trained ternary quantization.
\newblock {\em arXiv preprint arXiv:1612.01064}, 2016.

\bibitem[\protect\citeauthoryear{Zhuang \bgroup \em et al.\egroup
  }{2018}]{zhuang2018towards}
Bohan Zhuang, Chunhua Shen, Mingkui Tan, Lingqiao Liu, and Ian Reid.
\newblock Towards effective low-bitwidth convolutional neural networks.
\newblock In {\em Proceedings of the IEEE Conference on Computer Vision and
  Pattern Recognition}, pages 7920--7928, 2018.

\end{thebibliography}

\end{document}